# FRAME-LEVEL SPECAUGMENT FOR DEEP CONVOLUTIONAL NEURAL NETWORKS IN HYBRID ASR SYSTEMS

*Xinwei Li, Yuanyuan Zhang, Xiaodan Zhuang, Daben Liu*

Apple
{xinwei_li2, yuanyuan_zhang5, xiaodan_zhuang, daben_liu}@apple.com

## ABSTRACT

Inspired by SpecAugment — a data augmentation method for end-to-end ASR systems, we propose a frame-level SpecAugment method (f-SpecAugment) to improve the performance of deep convolutional neural networks (CNN) for hybrid HMM based ASR systems. Similar to the utterance level SpecAugment, f-SpecAugment performs three transformations: time warping, frequency masking, and time masking. Instead of applying the transformations at the utterance level, f-SpecAugment applies them to each convolution window independently during training. We demonstrate that f-SpecAugment is more effective than the utterance level SpecAugment for deep CNN based hybrid models. We evaluate the proposed f-SpecAugment on 50-layer Self-Normalizing Deep CNN (SNDCNN) acoustic models trained with up to 25000 hours of training data. We observe f-SpecAugment reduces WER by 0.5-4.5% relatively across different ASR tasks for four languages. As the benefits of augmentation techniques tend to diminish as training data size increases, the large scale training reported is important in understanding the effectiveness of f-SpecAugment. Our experiments demonstrate that even with 25k training data, f-SpecAugment is still effective. We also demonstrate that f-SpecAugment has benefits approximately equivalent to doubling the amount of training data for deep CNNs.

## 1. INTRODUCTION

Many deep neural network architectures have been proposed in recent years to improve Automatic Speech Recognition (ASR), such as deep convolutional neural network (CNN) models with over 50 convolutional layers [1, 2, 3] for hybrid ASR systems in which the deep neural networks combined with HMM are used as the acoustic models, and end-to-end models [4, 5] which integrate an acoustic model, a language model, and a pronunciation model into a single neural network. However, these models usually contain tens to hundreds of millions of parameters which tend to overfit even with thousands of hours of training data.

Data augmentation increases the amount or diversity of existing training data to alleviate the overfitting problem for deep neural networks. Many such methods have been proposed in the literature for ASR over the years. Vocal Track Length Perturbation (VTLP) is introduced in [6] to add variations to the speech data by applying the vocal track length normalization (VTLN) in a reverse manner. Speed Perturbation [7] increases the quantity of training data by changing the speed of the audio signal to produce three versions of the original audio with speed factors 0.9, 1.0 and 1.1. [8, 9] synthesize noisy speech by superposing the clean speech audio with additional noisy audio to improve the model robustness in noisy environments. An acoustic room simulator is explored in [10, 11] to improve ASR performance for far-field use cases in noisy environment. Synthesized speech generated from text by TTS systems is used in [12, 13, 14] to improve the performance of ASR systems. Most of these data augmentation methods for ASR operate on the audio level. The process is usually complicated and resource consuming especially for large scale datasets.

SpecAugment as a data augmentation method is shown to be effective in improving the performance of end-to-end Listen, Attend, and Spell (LAS) [5] ASR models for small to medium tasks in [15]. The effectiveness of SpecAugment for end-to-end models on large scale datasets is demonstrated in [16]. This data augmentation method avoids generating large amount of augmented audio data by directly operating on the log Mel filter bank features of the input audio online during training. SpecAugment has also been applied to BLSTM models to improve the hybrid ASR systems using small to medium scale datasets in [9, 17, 18]. The multistream hybrid ASR framework in [19, 20] applied an augmentation method called feature dropout which is similar to frequency masking to improve the robustness of the multistream ASR system.

The original SpecAugment was evaluated mostly on LSTM based ASR models while CNN also plays an important role for product-level ASR systems [3, 30, 31, 32]. In this paper, we propose a frame-level SpecAugment method (f-SpecAugment) to improve the performance of self-normalizing deep CNN (SNDCNN) models [3] for our CNN based hybrid ASR systems. Instead of applying the transformations at the utterance level as in [9, 15, 16, 17, 18], f-SpecAugment applies the transformations to each convolution window input to the CNN independently, which fits better to CNN compared with the original utterance-level SpecAug. f-SpecAugment can be applied to the mini-batch on-the-fly during training with less than 5% additional computational cost which makes it scalable to large training datasets.

Data augmentation tends to be more effective for small scale hybrid ASR systems trained with hundreds of hours of speech data than those with much larger training data. The improvement from SpecAugment for hybrid ASR systems in [9, 17, 18, 19, 20] are all achieved on small to medium scale datasets. It's interesting to see how it performs with large scale datasets. We evaluate f-SpecAugment for four languages using training data up to 25000 hours, on a 50-layer SNDCNN model for hybrid ASR systems. 0.5-4.5% relative word error rate (WER) reduction is observed from f-SpecAugment across different test sets. To our knowledge, this is the first work demonstrating success of SpecAugment on hybrid ASR models trained with large scale datasets. By controlling the amount of data used in the training, we demonstrate that tasks with different scales of training data can all benefit from f-SpecAugment.

## 2. ACOUSTIC MODEL

Self-normalizing deep CNNs (SNDCNN) [3] are used as the acoustic models in our experiments. As an enhanced variant of ResNet [1] with a deep CNN topology, SNDCNN addresses the problem of vanishing/exploding gradients by utilizing the scaled exponential linear unit (SELU) [21] as the activation function. The SELU activation function is formulated as Eq. (1).

$$\mathbf{selu}(x) = \lambda \begin{cases} x & \text{if } x > 0 \\ \alpha e^x - \alpha & \text{if } x \leq 0 \end{cases} \quad (1)$$

with $\alpha \approx 1.6733$ and $\lambda \approx 1.0507$.

With the self-normalizing properties of the SELU activation function, SNDCNN simplifies the model topology by removing the needs for shortcut connections and batch normalization as used in ResNet [3, 21].

Compared with ResNet, SNDCNN can achieve the same or lower WER with 60%-80% training and inference speedup for hybrid ASR systems [3].

## 3. FRAME-LEVEL SPECAUGMENT

SpecAugment was proposed as a data augmentation method for training LSTM based end-to-end ASR models in [15]. The augmentation policy includes three transformations: time warping, frequency masking, and time masking which are applied to the log Mel spectrograms of speech audio at the utterance level during training to improve the generalization performance of the end-to-end models.

Unlike the LSTM based end-to-end ASR models which are trained at utterance level, CNN based hybrid acoustic models are trained at frame level using the alignments between the feature frames and output labels. A context window around the current feature frame is fed as input to CNN models to predict the current label. Context windows are processed independently during training. We propose to apply SpecAugment at frame level to each context window for CNN based models. f-SpecAugment adopts the policy similar to the utterance level SpecAugment [15] with some modifications.

- Time warping with parameter $W$ applied via the function *sparse_image_warp* of *tensorflow*: time warping at the utterance level doesn't fit hybrid acoustic model training as mentioned in [9, 17], because it invalidates the alignments generated from the unwarped features. In contrast, by applying SpecAugment at frame level, we can avoid this problem by keeping the center frame fixed and applying time warping on one or both sides to make it applicable to hybrid acoustic models. A context window with $\tau$ frames of $D$-dimensional log Mel filter bank features is viewed as an image where the time axis is horizontal and the frequency axis is vertical. Suppose $D$ is even. To simplify the operation, we reshape the image into [$\tau$, 2, $D/2$], where $D/2$ is the number of channels, apply the warping function, and reshape it back to [$\tau$, $D$, 1] after the warping. A displacement $w$ is chosen from a uniform distribution from $-W$ to $W$, i.e., $w \sim U(-W, W)$. A random starting time step $w_0$ is uniformly chosen from the segment$[W, \lfloor \tau/2 \rfloor - W) \cup (\lfloor \tau/2 \rfloor + W, \tau - W]$. We exclude the segment $[\lfloor \tau/2 \rfloor - W, \lfloor \tau/2 \rfloor + W]$ to prevent frames from moving from one side of the center frame to the other. Two points $[w_0, 0]$ and $[w_0, 1]$ on the horizontal edges of the reshaped image are warped together either to the left or right by a distance $w$ with the center frame fixed. It's optional to fix the four corner points on the boundary.

- Frequency masking with parameter $F$: A mask size $f$ is chosen from a uniform distribution from 0 to $F$, i.e., $f \sim U(0, F)$. A starting point $f_0$ is chosen from $U(0, D - f)$ where $D$ is the dimension of the filter bank features. $f$ consecutive filter bank feature dimensions $[f_0, f_0 + f)$ are masked.

- Time masking with parameter $T$: A mask size $t$ is chosen from $U(0, T)$. A starting point $t_0$ is chosen from $U(0, \tau - t)$ where $\tau$ is the context window size. $t$ consecutive feature frames $[t_0, t_0 + t)$ are masked. Compared with the utterance level time masking, applying time masking at frame level to each context window independently can introduce more variations to the data for CNN based hybrid acoustic model training.

Multiple frequency and time masks can be applied to each context window in f-SpecAugment. For efficiency, all the context windows within a mini-batch share the same transformations. Zero is used as the masking value in our experiments. Because the log Mel filter bank features are globally normalized to have zero mean and unit variance, setting the masking value to zero is equivalent to setting it to the global mean value.

## 4. EXPERIMENTS

We evaluate the effectiveness of f-SpecAugment using a 50-layer SNDCNN model (SNDCNN-50) [3] for hybrid ASR systems.

### 4.1 Training and test data

The internal anonymized mobile device virtual assistant and dictation datasets for four languages (Japanese, French, Indian English and Mandarin) are used in the experiments. Training data contains mostly 16kHz wide band audio mixed with 5% to 10% 8kHz narrow band audio. 8kHz audio is upsampled to 16kHz before feature extraction during training and testing. We report the WERs mainly on two tasks with 16kHz test data: an assistant task and a dictation task. Table 1 shows the amount of training and 16kHz test data for each language. For Japanese, we also report the WERs for models on an 8kHz assistant test set with 8 hours of audio.

Table 1: *amounts of data for languages*

| Language | Training | Test (16kHz) | |
|---|---|---|---|
| | | Assistant | Dictation |
| Japanese | 9.6k hours | 39 hours | 35 hours |
| French | 7.7k hours | 28 hours | 33 hours |
| Indian English | 6.3k hours | 36 hours | 20 hours |
| Mandarin | 25k hours | 35 hours | 35 hours |

### 4.2 Baseline model

SNDCNN-50 models trained without SpecAugment serve as the baseline for our experiments. As reported in Section 2, the topology of SNDCNN-50 is revised from the 50-layer ResNet (ResNet-50) [1]. Acoustic features are 80-dimensional log Mel filter bank coefficients extracted with 25ms window size and 10ms frame shift. A context window of 41 frames, consisting of a single center frame together with 20 frames on the left and 20 frames on the right, is used as input to SNDCNN-50. All models are first trained with the cross entropy objective function followed by the sequence discriminative training with the MMI objective function [22, 23, 24, 33, 34, 35]. Training is parallelized with 32 GPUs using Blockwise Model-Update Filtering (BMUF) [25] algorithm. For cross entropy training, we adopt the Newbob learning rate scheduling as implemented in the Quicknet software [29] with an initial learning rate 0.008 and a decay factor of 0.5 in all experiments. The maximum number of training epochs is set to 60. Training is early stopped if there's no frame accuracy improvement on the validation set for two consecutive epochs. MMI training is performed with a constant learning rate 7.0e-5 for a fixed number of training steps. After MMI training, the model with the best WER on a development set is chosen for evaluation on the test sets. All the tests are performed with 4-gram language models.

### 4.3 Frame level SpecAugment configurations

f-SpecAugment is applied to each context window represented with a 41 frames by 80 filterbanks log Mel spectrogram. We picked up the following augmentation parameters experientially without elaborate tuning. Further improvement can be expected through fine-tuning the augmentation parameters.

- Time warping parameter: $W = 5$
- Frequency masking with 1 mask: $F = 15$
- Time masking with 1 mask: $T = 10$

f-SpecAugment is applied to both cross entropy training and MMI training. Lattices for MMI training are generated with the original features without f-SpecAugment. We didn't do any adjustment to the learning rate scheduling for model training with f-SpecAugment. Using the same learning rate scheduling as the baseline model training, we didn't observe any issues in convergence.

### 4.4 Comparison of augmentation transformations

To investigate the effectiveness of the three transformations used in f-SpecAugment, we compared their contributions in Table 2. All models in the table were trained with the same 1k hours of audio randomly sampled from the Japanese training set. We first evaluated the models on the Japanese 16kHz assistant test set. To compare the f-SpecAugment with utterance level SpecAugment, we include the WER of a model trained with utterance level SpecAugment (u-SpecAug) in Table 2. The utterance level SpecAugment adopts one frequency mask with $F = 15$ and one time mask with $T = 30$ which gives us the best performance on the test set.

All training experiments with or without SpecAugment converged in roughly the same number of epochs. Given the low computational cost of the transformations used for the augmentation, there's less than 5% increase in training time. Specifically, the experiments without time warping had almost no training time increase. Time warping is more computationally expensive than time masking and frequency masking.

Table 2: *Comparison of augmentation transformations (T: time, F: frequency)*

| | Transformations | WER (%) | WERR (%) |
|---|---|---|---|
| **16kHz assistant test set** | | | |
| Baseline | None | **6.45** | |
| u-SpecAug | T+F-masking | **6.13** | 5.0 |
| f-SpecAug | T-warping | **6.30** | 2.3 |
| | F-masking | **6.22** | 3.6 |
| | T-masking | **6.26** | 2.9 |
| | T+F-masking | **6.01** | 6.8 |
| | All three | **6.01** | 6.8 |
| **8kHz assistant test set** | | | |
| Baseline | None | **10.70** | |
| f-SpecAug | T+F-masking | **9.65** | 9.8 |
| | All three | **9.31** | 13.0 |

Table 3: *WERs on 16kHz assistant test sets*

| Model | WER (%) Baseline | WER (%) f-SpecAug | WERR (%) |
|---|---|---|---|
| Japanese-1 | 5.59 | **5.38** | 3.8 |
| Japanese-2 | 5.59 | **5.38** | 3.8 |
| French | 6.20 | **5.92** | 4.5 |
| Indian English | 6.74 | **6.64** | 1.5 |
| Mandarin | 5.25 | **5.15** | 1.9 |

Table 4: *WERs on 16kHz dictation test sets*

| Model | WER (%) Baseline | WER (%) f-SpecAug | WERR (%) |
|---|---|---|---|
| Japanese-1 | 4.72 | **4.61** | 2.3 |
| Japanese-2 | 4.72 | **4.59** | 2.8 |
| French | 6.54 | **6.32** | 3.4 |
| Indian English | 10.43 | **10.02** | 3.9 |
| Mandarin | 6.36 | **6.33** | 0.5 |

Table 5: *WERs on 8kHz Japanese assistant test set*

| Model | WER (%) | WERR (%) |
|---|---|---|
| Baseline | **8.67** | |
| Japanese-1 | **8.52** | 1.7 |
| Japanese-2 | **8.09** | 6.7 |

Table 2 shows that among the three transformations in f-SpecAugment, when applied separately in the training, frequency masking brings the most improvement while time warping brings the least. Further improvement is achieved when we combine time masking and frequency masking.

f-SpecAugment with time and frequency masking outperforms the utterance level SpecAugment with time and frequency masking by about 2% relatively for CNN based hybrid models. Because the overall effect of frequency masking is similar between frame level and utterance level SpecAugment, the improvement is mainly attributed to the more variations introduced to the data by applying the time masking at the frame level. Time masking in utterance level SpecAugment may be improved by using multiple time masks or adaptive time masking policy [16], in which the number of time masks and/or the size of the time mask vary depending on the length of the input. But it introduces more parameters to tune. Similarly, we may fine-tune the number of time masks and the size of the time mask for f-SpecAugment to improve its performance. Overall, f-SpecAug is more flexible and fits better to the CNN based hybrid model training. Combining time warping with time and frequency masking doesn't bring further improvement on the 16kHz assistant test set, which is consistent with the observation in [15].

In the bottom part of Table 2, we further evaluate the baseline model, f-SpecAugment models with and without time warping, on the 8kHz assistant test set. The 8kHz audio is upsampled to 16kHz during recognition. The last two rows show that time warping brings additional 3.5% WER reduction on top of time and frequency masking on the 8kHz test data, which is less represented in the training set. It seems that time warping can reduce the mismatch between the two channels in training data, and benefits the one with far less training data more. This is also validated in the larger experiments below.

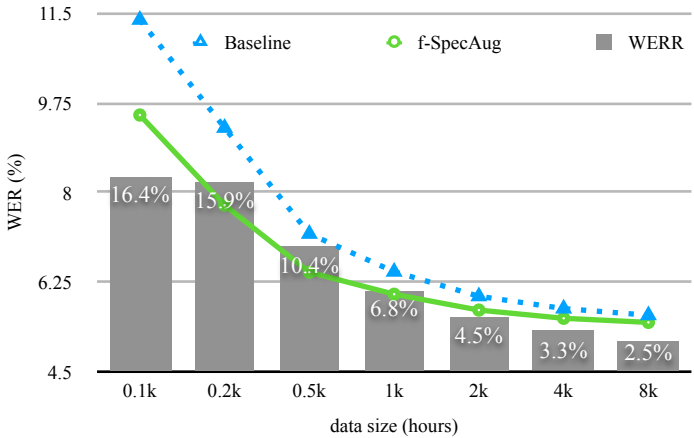

Figure 1: *Performance of f-SpecAug with different data sizes*

### 4.5 Results on large scale datasets

To study the effectiveness of f-SpecAugment on large scale datasets, we conducted experiments on the full training sets for Japanese, French, Indian English and Mandarin. We trained two Japanese models with f-SpecAugment on the full 9.6k hours training dataset. One was trained with only time and frequency masking (Japanese-1). The other was trained with all three transformations (Japanese-2). For the full French, Indian English and Mandarin training datasets, f-SpecAugment was applied with only time and frequency masking to model training. The WERs on the 16kHz assistant test sets for the four languages are presented in Table 3. The WERs on the 16kHz dictation test sets are presented in Table 4. From the two tables, it can be seen that the models trained with f-SpecAugment consistently outperform the baseline models for all four languages on both test sets. The relative WER reduction ranges from 0.5% to 4.5%. Even with 25k hours training set, f-SpecAugment still achieves 1.9% relative WER reduction on Mandarin assistant test set. Although the gain is not as large as reported for end-to-end systems [15, 16], it's still significant for hybrid ASR systems trained with up to 25000 hours of training data. If we compare the two Japanese models, little gain is observed from time warping on the two 16kHz test sets when it's combined with the other two transformations.

The WERs on the 8kHz assistant test sets for the Japanese models are presented in Table 5. Similar to the observation on the models trained with 1k hours of data, large improvement is achieved on the 8kHz test data from time warping on top of time and frequency masking. The model trained with time warping (Japanese-2) outperforms the model trained without time warping (Japanese-1) by 5% relatively.

### 4.6 Effect of training data size

To further study the effectiveness of f-SpecAugment for hybrid ASR systems with different dataset sizes, we conducted a set of controlled experiments by training models with subsets of the Japanese dataset. The subsets are created by randomly selecting 100 hours, 200 hours, 500 hours, 1k hours, 2k hours, 4k hours, and 8k hours of training data from the whole dataset. Smaller subsets are always included in the larger subsets. For each subset, we trained a baseline model without f-SpecAugment and a model with f-SpecAugment.

The WER results on the Japanese 16kHz assistant test set for the fourteen models are plotted in Figure 1. The x-axis represents the amount of training data in hours used for the training. The y-axis represents the WER on Japanese assistant test set. The two curves show how WER evolves with the increase of training data. The dotted curve with triangle marks is for the baseline models. The solid curve with circle marks is for the models trained with f-SpecAugment. If we compare the two curves vertically, we see that models trained with f-SpecAugment consistently outperform the baseline models for all dataset sizes. The relative WER reduction decreases with the increase of the training data as presented in the grey bars which suggests that the model suffers more from overtraining with smaller training sets. Comparing the two curves horizontally, we can see that the improvement from f-SpecAugment is approximately equivalent to doubling the training data.

## 5. CONCLUSIONS

We propose a frame-level SpecAugment (f-SpecAugment) method to improve the performance of deep CNN models for hybrid HMM based ASR systems. Similar to the original utterance level SpecAugment proposed for end-to-end ASR systems, f-SpecAugment involves three transformations: time warping, frequency masking, and time masking. However, they are applied to each convolution window independently instead of to the whole utterance. We first show that f-SpecAugment with more variations in frame level time masking outperforms utterance level SpecAugment by 2% using a small dataset. Then, we demonstrate the effectiveness of f-SpecAugment for large scale hybrid ASR systems by evaluating it on SNDCNN-50 models using internal datasets from four languages with 6k to 25k hours of training data. On the two 16kHz test sets for all three languages, applying f-SpecAugment in training results in 0.5% to 4.5% relative WER reductions. Although time warping, in addition to time and frequency masking, doesn't further improve accuracy on the 16kHz test sets, it results in 5% relative WER reduction on the Japanese 8kHz test set. Similar improvement is observed on the models trained with only cross entropy loss without sequence training. The improvement comes with no increase of model complexity, no increase of training data and less than 5% increase in training time. Our controlled experiments further demonstrate that hybrid ASR models trained with a wide range of data scales benefit from f-SpecAugment. The results also show that the improvement from f-SpecAugment is approximately equivalent to doubling the training data. Next step, we plan to investigate the best strategies to combine SpecAugment with other data augmentation methods to improve ASR systems.

## 6. ACKNOWLEDGEMENTS

The authors would like to thank Kyuyeon Hwang, Zhen Huang, Xiaoqiang Xiao, Mu Su, Jun Zhang, Bing Zhang, and Donald McAllaster for valuable discussions.